\documentclass[conference]{IEEEtran}

\usepackage{cite} % for numeric citations
\usepackage{amsmath,amssymb,amsfonts} % for math equations, symbols and fonts
\usepackage{algorithmic} % for writing algorithms
\usepackage{graphicx} % for inserting pictures
\usepackage{textcomp} % extra text symbols
\usepackage{booktabs}

\usepackage{color}
\usepackage{enumitem}
\usepackage[dvipsnames]{xcolor}
\usepackage[hyphens]{url}
\usepackage[hyperfootnotes=false]{hyperref}
\usepackage{threeparttable}

\begin{document}

% \title{Fine-tuning Automatic Speech Recognition (ASR) for Accent-Specific Air Traffic Control Communications}
% \title{Fine-Tuning ASR for Accent-Specific ATC Communications}
\title{Adapting Automatic Speech Recognition for Accented Air Traffic Control Communications}

% \title{Analyzing Asian Accents in Air Traffic Operations}
%\title{Small-scale Automatic Speech Recognition Models for Accent-Specific Air Traffic Control Speech}
% \title{Automatic Speech Recognition for Accented Air Traffic Control Communications}
% MW: How about something like this
% \title{Accented ATC Speech ASR through Accent-Specific Fine-Tuning}
% Optimizing ASR for Accented ATC Speech through Accent-Specific Fine-Tuning
% \title{Analyzing Secondary-Spoken Accents in Air Traffic Operations}
% yz: let me think harder
% \title{Analyzing Asian Code Talking in Air Traffic Communications}
% \title{Analyzing Asian Code Talking Performance in Air Traffic Communications using Automated Speech Recognition Models}

\author{
    \IEEEauthorblockN{Marcus Yu Zhe Wee, Justin Juin Hng Wong, 
    Lynus Lim, Joe Yu Wei Tan, Prannaya Gupta, \\ Dillion Lim, En Hao Tew,
    Aloysius Keng Siew Han, Yong Zhi Lim}
    \IEEEauthorblockA{Air Emerging Technologies High-Speed Experimentations and Research (AETHER), \\ RSAF Agile Innovation Digital (RAiD), Republic of Singapore Air Force, Singapore}
    \{marcus\_yu\_zhe\_wee, justin\_wong\_juin\_hng, lynus\_lim, joe\_tan\_yu\_wei, prannaya\_gupta, \\ dillion\_lim, tew\_en\_hao, han\_keng\_siew\_aloysius, lim\_yong\_zhi1\}@defence.gov.sg

% icmt2025 double-blind
% \IEEEauthorblockN{Anonymous Authors}
}

\maketitle

\begin{abstract}

% Wait south east is not correct technically oops
Effective communication in Air Traffic Control (ATC) is critical to maintaining aviation safety, yet the challenges posed by accented English remain largely unaddressed in Automatic Speech Recognition (ASR) systems. Existing models struggle with transcription accuracy for Southeast Asian-accented (SEA-accented) speech, particularly in noisy ATC environments. This study presents the development of ASR models fine-tuned specifically for Southeast Asian accents using a newly created dataset. Our research achieves significant improvements, achieving a Word Error Rate (WER) of 0.0982 or 9.82\% on SEA-accented ATC speech. Additionally, the paper highlights the importance of region-specific datasets and accent-focused training, offering a pathway for deploying ASR systems in resource-constrained military operations. The findings emphasize the need for noise-robust training techniques and region-specific datasets to improve transcription accuracy for non-Western accents in ATC communications.
\end{abstract}

% yz: please avoid the use of \\ in scientific writing
%%%%%%%%%%%%%%%%%%%%%%%%%%%%%%%%%%%%%%%%%%%%%%%%%%%%%%%%%%%%%%%%%%%%%%%%%%%%%%%%%%%%%%%%%%%%%%%%%%%%%%%%%%%%%%%%%%%%%%%%%%
% INTRODUCTION
%%%%%%%%%%%%%%%%%%%%%%%%%%%%%%%%%%%%%%%%%%%%%%%%%%%%%%%%%%%%%%%%%%%%%%%%%%%%%%%%%%%%%%%%%%%%%%%%%%%%%%%%%%%%%%%%%%%%%%%%%%
\section{Introduction} \label{introduction}

Air Traffic Control (ATC) plays an indispensible role in ensuring the safety and efficiency of global aviation. Effective communication between Air Traffic Controllers (ATCOs) and pilots is critical as it directly impacts the operational safety of flight missions in and out of the military. The International Civil Aviation Organization (ICAO) mandates the use of standardized phraseology to minimize communication ambiguities \cite{Ohneiser2021IntegratingEA}. However, real-world ATC communications may deviate from these standards, especially in multilingual regions where English is spoken with various non-native accents \cite{Fan2023EnhancingMS}. These deviations, coupled with the inherent complexity of ATC interactions, pose significant challenges for Automatic Speech Recognition (ASR) systems tasked with transcribing ATC communications for operational use.

Despite substantial advancements in ASR technologies, modern models struggle with domain-specific challenges unique to ATC environments \cite{Zuluaga2023LessonsLI}. These challenges include rapid speech rates, overlapping conversations, and high levels of ambient noise, such as radio interference. Moreover, Southeast Asian-accented (SEA-accented) English remains severely underrepresented in public datasets, resulting in suboptimal transcription accuracy for this region's ATC communications. Such deficiencies are particularly concerning for aviation safety, as errors in transcriptions can lead to severe consequences, including flight delays, near-misses, or even accidents.

The introduction of models like OpenAI's \emph{Whisper} has brought new opportunities for robust ASR solutions. Whisper has demonstrated remarkable robustness across various accents and noise conditions in general speech datasets \cite{radford2023robust}. However, it has shown limited success in adapting to specialized domains, such as ATC, where region-specific accents and terminologies are prevalent. This limitation highlights the need for fine-tuning ASR models using datasets that capture the nuances of SEA-accented ATC speech.

In this study, we address the gaps in ASR performance for SEA-accented ATC speech by introducing an accent-specific fine-tuning approach. We leverage a newly developed SEA-accented ATC dataset and employ noise-resilient training strategies tailored to this domain. Our research focuses on three primary objectives: 
% \begin{itemize}
\begin{enumerate}
    \item Enhancing transcription accuracy for SEA-accented English;
    \item Improving robustness to noisy ATC environments;
    \item Ensuring computational efficiency to enable deployment on resource-constrained hardware, such as those used in military operations.
\end{enumerate}
% \end{itemize}

To achieve these objectives, a rigorous evaluation of our fine-tuned Whisper models is conducted, comparing to baseline and existing state-of-the-art models. Our findings demonstrate better performance in Word Error Rate (WER) of \textbf{0.0982} or 9.82\% on our dataset, underscoring the effectiveness of region-specific fine-tuning. We also explore the broader implications of our approach for both civilian and military ATC contexts, where accurate and efficient transcription of ATC communications is critical for operational success.

The flow of the paper is as follows: Section~\ref{militaryrelevance} examines the relevance of ASR in military ATC contexts, highlighting key operational benefits and challenges. Section~\ref{litreview} provides a detailed review of existing ASR research for ATC, with a focus on accented speech and noise robustness. Section~\ref{methodology} outlines our research objectives and details the dataset creation process and the fine-tuning pipeline. Experimental results are presented and analyzed in Section~\ref{results}, followed by a discussion of key insights and broader implications. Finally, Section~\ref{conclusion} summarizes our contributions, and Section~\ref{futurework} outlines future research directions aimed at further advancing ASR systems for accented ATC speech.

%introduction militaryrelevance litreview researchobj methodology results conclusion future work

% \ref{researchobj}
%%%%%%%%%%%%%%%%%%%%%%%%%%%%%%%%%%%%%%%%%%%%%%%%%%%%%%%%%%%%%%%%%%%%%%%%%%%%%%%%%%%%%%%%%%%%%%%%%%%%%%%%%%%%%%%%%%%%%%%%%%
% MILITARY RELEVANCE
%%%%%%%%%%%%%%%%%%%%%%%%%%%%%%%%%%%%%%%%%%%%%%%%%%%%%%%%%%%%%%%%%%%%%%%%%%%%%%%%%%%%%%%%%%%%%%%%%%%%%%%%%%%%%%%%%%%%%%%%%%
\section{Military Relevance} \label{militaryrelevance}
%% refined draft 2
Air Traffic Controllers (ATCOs) play a key role in air operations as they manage complex airspaces to ensure aircraft safety \cite{Smith2020UnderstandingRA}. In time-sensitive military operations where time is tight, ATCOs play an especially critical role in enabling the smooth execution and achieving mission success. 

% dillion: can look at this air force doctrine from USAF page 20, see if you want to use this as a reference \href{https://www.doctrine.af.mil/Portals/61/documents/AFDP_3-/3-52-AFDP-AIRSPACE-CONTROL.pdf}{link}

Traditionally, ATC speech transcriptions are obtained manually, requiring human ATCOs to listen to recordings and transcribe them into text. This process is time-consuming, labor-intensive, and prone to human error \cite{Zuluaga2020ATC}. Integrating Automatic Speech Recognition (ASR) systems into this workflow offers significant advantages, including enhanced command and control capabilities, reduced operational workload, and improved mission efficiency \cite{Wang2024ASR}.

% Below, 
We outline the 3 key military benefits of ASR integration:

\subsubsection*{\textbf{Reduced ATC Errors}}

ATC communication errors may have severe consequences in military operations, potentially leading to mission failure or accidents. Some communication errors include failure to challenge incorrect readbacks, incorrect call signs, use of non-standard phraseology, and missing or clipped call signs \cite{moon2011air}. These miscommunications can cause severe consequences and accidents, especially critical in military contexts, where operational precision can influence the outcome of international conflicts.

The use of ASR technology in transcription systems has the potential to provide accurate, real-time transcriptions of ATC communications, serving as a reliable reference for ATCOs. This could reduce communication errors and enhance operational safety, contributing to the seamless execution of military missions.

\subsubsection*{\textbf{Enhanced ATC Surveillance}}

Military organizations also engage in Intelligence, Surveillance, and Reconnaissance (ISR) operations, generating vast amounts of data requiring continuous analysis. Specialized personnel may have to manage and tune into various streams of ATC communication channels while facing limitations in manpower and cognitive capacity. The adoption of ASR systems could automate the transcription of audio streams into text, significantly reducing the workload. This allows surveillance teams to visually monitor multiple ATC communication channels, reducing cognitive load and aiding comprehension and memory \cite{Zdorova2021DoWR}. Moreover, accent-specific ASR systems alleviate reliance on scarce language experts, minimizing bottlenecks in the data processing pipeline and enabling analysts to focus on higher-value intelligence tasks.

% \subsubsection*{\textbf{Systems Enhancement}}
% \textcolor{red}{This part we could probably remove if exceed word count}
% Speech recognition technologies, such as AcListant~\cite{rataj2019aclistant}, demonstrate how automation can improve ATCO efficiency by streamlining routine tasks. These tools reduce reliance on manual inputs like keyboard and mouse interactions, allowing ATCOs to concentrate on critical decision-making. The integration of ASR systems into ATC workflows enhances situational awareness, reduces cognitive demands, and ensures faster, more accurate responses in high-pressure scenarios.

\subsubsection*{\textbf{Enhanced Training and Incident Investigation}}

Speech recognition can be used for training by evaluating the accuracy of the phraseology used by trainees without the need for a pseudo-pilot while doing simulator training ~\cite{kopald2013applying}. In addition, a clear, searchable record of communications, the transcribed text from ASR systems can support investigations into communications that lead up to incidents such as near-misses or to investigate protocol breaches.

% Im not sure if this needs to cite as its 
% ~\cite{see comment 1 below} 
% since Cockpit Voice Recorder (CVR) recordings are typically not made available ~\cite{see comment 2 below}, leaving accurate CVR transcriptions as an important source of information. 
% Comment 1: https://www.faa.gov/air_traffic/publications/atpubs/foa_html/chap3_section_4.html this as ref? idk
% Recorders may be used to monitor any position for evaluation, training, or quality control purposes.

% Comment 2: https://news.erau.edu/-/media/files/news/hall-kapustin-paper-2019.pdf this one probably
%%%%%%%%%%%%%%%%%%%%%%%%%%%%%%%%%%%%%%%%%%%%%%%%%%%%%%%%%%%%%%%%%%%%%%%%%%%%%%%%%%%%%%%%%%%%%%%%%%%%%%%%%%%%%%%%%%%%%%%%%%
% LITERATURE REVIEW %
%%%%%%%%%%%%%%%%%%%%%%%%%%%%%%%%%%%%%%%%%%%%%%%%%%%%%%%%%%%%%%%%%%%%%%%%%%%%%%%%%%%%%%%%%%%%%%%%%%%%%%%%%%%%%%%%%%%%%%%%%%
\section{Literature Review} \label{litreview}

Studies have been conducted on applying Automatic Speech Recognition (ASR) on Air Traffic Control (ATC) communications, as well as ASR on accented speech. However, few studies focus on the combined problem of accented speech, within the domain of ATC. The literature review of existing datasets, proposed models, and projects is below.
%-------------------------------------------------------------------------------------------------------------------------
% Available Public Datasets
%-------------------------------------------------------------------------------------------------------------------------
\subsection{Public ATC Speech Datasets}

Many research projects have collected ATC speech datasets that are used in research today. The prominent ATC Speech datasets that are commonly used in research and their featured languages and accents are depicted in Table \ref{table:1} \cite{Wang2024EnhancingAT}. 

\begin{table}[htbp]
\centering
\caption{Public ATC Transcription Datasets Available}

\begin{threeparttable}
\begin{tabular}{|l|c|p{0.5\linewidth}|}
    \hline
    \textbf{Dataset Name} & \textbf{Data Size} & \textbf{Language and Accent} \\
    \hline
    LDC94S14A~\cite{LDC94S14A} & 72.5h & US English\\
    \hline
    N4 NATO & 72.5h & Canadian, German, Dutch and British-accented English\\
    \hline
    HIWIRE~\cite{segura2007hiwire} & 28.3h & French, Greek, Italian and Spanish-accented English \\
    \hline
    Madrid Airport & 11.8h & Spanish, English \\
    \hline
    Madrid ACC & 100h & Spanish, English \\
    \hline
    ATCOSIM~\cite{hofbauer2008atcosim} & 10.7h & German, Swiss, and French accented English \\
    \hline
    AcListant & 8h & German and Czech-accented English \\
    \hline
    DataComm & 120h & US English \\
    \hline
    ATCSC & 4800u* & English \\
    \hline
    AIRBUS-ATC & 59h & French-accented English \\
    \hline
    MALORCA & 10.9h & German and Czech-accented English \\
    \hline
    UWB-ATCC & 179h & Czech-accented English\\
    \hline
    SOL-Twr & 1993u* & Lithuanian-accented English \\
    \hline
    SOL-Cnt & 800u* & German-accented English \\
    \hline
    HAAWAII & 34h & Icelandic and British-accented English \\
    \hline
    ATCO$^2$~\cite{zuluaga2023atco2} & 5285h & English \\
    \hline
    ATCSpeech~\cite{yang2019atcspeech} & 58h & Mandarin Chinese, English \\
    \hline
\end{tabular}
\begin{tablenotes}
\item *utterances
\end{tablenotes}
\end{threeparttable}
\label{table:1}
\end{table}

% https://www.semanticscholar.org/reader/c38b2029f5aa484accc5feeeda3658dbe9d3cc41 refer to this

Notable corpora include ATCOSIM\cite{hofbauer-etal-2008-atcosim}, the first publicly available civil aviation ATC corpus, featuring 10 hours of utterances by non-native English speakers recorded in simulated environments. The AcListant project\cite{rataj2019aclistant} created a dataset with recordings from three European controllers using ICAO-standardized phraseologies. Similarly, the AIRBUS-ATC\cite{pellegrini2018airbus} corpus targets non-native speech and poor audio quality, often serving as a benchmark in ASR research.

The MALORCA project\footnote{\url{https://www.malorca-project.de/}} provided high-quality datasets with semi-supervised learning methods for unlabeled data, while UWB-ATCC\cite{smidl2019ATCC} offers annotated ATC recordings from Czech airspace. The SOL-Twr and SOL-Cnt datasets from SESAR projects\cite{Wang2024EnhancingAT} capture communications in Lithuanian and Viennese airspaces. The HAAWAII project\cite{helmke2023haawaii} includes 19 hours of annotated London approach data and 15 hours of Icelandic en-route communications.

The ATCO2 project\cite{zuluaga2023atco2} constructed the largest ATC corpus to date, with over 5281 hours of automatically transcribed recordings from 10 international airports. Meanwhile, ATCSpeech\cite{yang2019atcspeech} is notable for including accented Mandarin Chinese and English utterances, tailored for multilingual ATC communications. Finally, LiveATC, an online platform, provides an extensive archive of real-time ATC recordings, which can be manually transcribed or processed with semi-supervised methods.

The corpora above mainly feature Western languages and Western-accented English as their training data. Because studies have shown that \emph{Whisper}-based ASR models are unable to effectively generalize to new accents absent from their training data \cite{Pan2024AssessmentAA}, the lack of Southeast Asian-accented (SEA-accented) English means that current ATC models available that are trained on them will not perform as well on SEA-accented English. This is also reflected in our model tests in Table \ref{table:werresults}.

The implications of the lack of accent representation mean that current models trained on these datasets may not be suitable for military organizations coming from countries that do not speak western-accented English.

\subsection{ASR Metrics}
% \textcolor{red}{yz: wer is nice to see and use but no reference} \\
Word error rate (WER) is a commonly used metric in model selection and to benchmark ASR model performance \cite{Kheddar2023DeepTL}. This metric is used in most of the papers \cite{Zuluaga2023LessonsLI, Wang2024EnhancingAT, Ahrenhold2023ValidatingAS, Iwamoto2022HowBA, Prabhu2023AccentedSR,li2021accent,Deng2021ImprovingAI, jahchan2021towards, Jelassi2024RevolutionizingRA, Graham2024EvaluatingOW, Sanabria2023TheEI, pan2024assessment, han2021supervised, aksenova2022AccentedSR, Maison2023ImprovingAS}, and also used by \emph{Whisper} \cite{RadfordWhisper2022}. In our subsequent evaluation and tests conducted, Combined WER is used, calculating the WER from the combined predicted transcriptions and ground truths across the entire dataset. The formula for Combined WER used is as such:

\begin{equation*}
    \text{Combined WER} = \frac{S + D + I}{N}
\end{equation*}

where:
\begin{itemize}
    \item \( S \) = Total number of substitutions (words incorrectly replaced by other words),
    \item \( D \) = Total number of deletions (words missing in the recognized output),
    \item \( I \) = Total number of insertions (extra words added in the recognized output),
    \item \( N \) = Total number of words in the reference (ground truth).
\end{itemize}

%-------------------------------------------------------------------------------------------------------------------------
% Public ATC Speech Transcription Models
%-------------------------------------------------------------------------------------------------------------------------
\subsection{ATC Speech Recognition}
% Radio Noise
ASR speech transcription models perform to high degrees of accuracy in benchmarks for state-of-the-art speech datasets\footnote{\url{https://paperswithcode.com/sota/automatic-speech-recognition-on-lrs2}}. Studies show that ASR models still struggle to transcribe ATC speech due to the presence of large amounts of noise and high speech rates \cite{Lin2021SpokenIU,Zuluaga2023LessonsLI} as well as ATC-specific language ontologies \cite{Ahrenhold2023ValidatingAS}. 

\subsubsection{ATC Speech Rate} Air Traffic Controllers (ATCOs) have to relay information to each other quickly due to the high-speed nature of air traffic, which reduces the intelligibility of spoken utterances, causing ASR models trained on regular speech to be unable to transcribe effectively ~\cite{Wang2024EnhancingAT}.

\subsubsection{Radio Noise} The ATC environment is characterized by complex background noise, including environmental noise, device noise, radio transmission noise, and speech echo \cite{Yu2023ROSEAR}. The speech echo, in particular, is a specific overlapping phenomenon generated by the ATC communication between the sent and received ATCO speech. Another component of noise, radio transmission noise, is the result of ATC being transmitted over High Frequency (HF), Very High Frequency (VHF) or Ultra High Frequency (UHF) bandwidths that are susceptible to noise caused by static, radio frequency interference, or thermal noise \cite{Lin2021SpokenIU}. Auditory information is encoded using amplitude modulation (AM) because it is less susceptible to the capture effect than frequency modulation (FM)~\cite{zeek1949investigation}. However, AM signals are also less robust against noise and interference, which has been shown to degrade the performance of ASR models~\cite{fritz2024analyzing}.

Due to the complex amalgamation of noises, ATC speech often sounds muffled and unintelligible with a low signal-to-noise ratio (SNR). This poses a problem to ASR systems as the noise will either have to be learned by the ASR model or removed through de-noising measures in pre-processing. However, studies have also shown that the use of de-noisers and enhancers could potentially also result in degraded performance in ASR models, making ASR transcription a tricky problem to address \cite{Iwamoto2022HowBA}.

\subsubsection{ATC-Specific Language Ontologies}

ATC Communications has many region-specific phraseologies and language ontologies~\cite{Chen2023EffectsOL,Ahrenhold2023ValidatingAS}. ATC-specific vocabulary could include landmark names, callsigns, and keywords that ATCOs use to refer to protocols and flight actions. 

Studies have shown that \emph{Whisper} Models~\cite{radford2023robust} far outperform other ASR models in robustness towards transcribing domain-specific terminologies from diverse speech recognition contexts, seen in healthcare radiology transcription~\cite{Jelassi2024RevolutionizingRA}. Hence, we selected \emph{Whisper} as our candidate model for research.

\subsubsection{Available Whisper Models}

Before \emph{Whisper}, many ASR models were researched in the ATC field with varying degrees of accuracy. Most notably, Convolutional Neural Networks chained with factorized Time Delayed Neural Networks (CNN-TDNNF) were shown to perform well, achieving approximately 5.0\% WER for the ATCOSim dataset.

Subsequently, with the introduction of \emph{Whisper}, the Huggingface platform\footnote{\url{https://huggingface.co/}} has featured many community-trained \emph{Whisper} models for ATC data. Most notably, the \emph{Whisper-ATC}~\cite{jlvdoorn2024WhisperATC} project fine-tuned \emph{Whisper} ASR models for Western-accented ATC datasets and achieved significant results. \emph{Whisper-ATC} fine-tuned \emph{Whisper} models on ATCO2 and ATCOSim, beating the prior model with only 1.19\% WER on the ATCOSim dataset. 

%-------------------------------------------------------------------------------------------------------------------------
% Accented ATC
%-------------------------------------------------------------------------------------------------------------------------
\subsection{Study on Accents in ATC ASR}

\subsubsection{Accent Analysis in ASR}
Although studies show that accents affect the performance of ASR models due to the variations in pronunciation, intonation, and speech patterns \cite{Prabhu2023AccentedSR,Graham2024EvaluatingOW}, the impact of accents is less researched in the domain of ATC speech transcription.

Traditional analysis of accents shows that accents are complex as every individual has a unique accent, making it difficult to classify ~\cite{Sanabria2023TheEI}. Despite the difficulties in accent classification, studies show that people in the same geographical region usually share common accent features while having different accents \cite{Xie2024FromFE}. 

The main study conducted on the impact of accents on ATC transcriptions is a performance survey on how pre-trained models, such as \emph{Whisper} and other architectures, inferred on different accents of ATC Speech data causes transcription accuracy differences of up to 26.37\%~\cite{pan2024assessment}. Studies on how models fine-tuned to transcribe certain accents of ATC perform on other accents have yet to be explored.

This is relevant to military organizations because studies have shown that military organizations internally share accent features and speech styles that are reflective of their unique cultural and historical contexts \cite{Romanov2021TheUM}. This could pose further issues in the domain of accent-specific ATC speech transcription.

\subsubsection{Accent Fine-Tuning}

Accent variability remains a significant challenge in ASR systems, especially for non-Air Traffic Control (ATC) accented speech. Various approaches have been explored to enhance ASR performance under such conditions. These include accent-aware techniques~\cite{Prabhu2023AccentedSR,li2021accent}, accent-agnostic methods~\cite{sun2018domain,han2021supervised}, and accent adaptation using both supervised and unsupervised strategies~\cite{Deng2021ImprovingAI,Prabhu2023AccentedSR}. 

For ATC ASR, Airbus proposed a HMM-TDNN-based accent-agnostic approach~\cite{jahchan2021towards}. However, this method exhibited catastrophic forgetting, as the Word Error Rate (WER) for previously seen accents increased when fine-tuning the model with Chinese-accented data.

A promising solution for accent-specific fine-tuning involves constructing datasets with sufficient accent representation. Such datasets have demonstrated the potential to significantly improve ASR performance, even when using limited real accented data during supervised fine-tuning~\cite{aksenova2022AccentedSR}. Additionally, region-specific fine-tuning for accent adaptation in general speech has shown notable improvements in accuracy for targeted accents~\cite{Maison2023ImprovingAS}.

Building upon these findings, we hypothesize that combining region-specific fine-tuning with robust accent representation in the training data can effectively mitigate the challenges of accent variability, particularly in ATC ASR contexts.

\section{Methodology} \label{methodology}

Our research methodology is designed to address the challenges of developing robust ASR systems for Southeast Asian-accented (SEA-accented) Air Traffic Control (ATC) speech. 
% The key research objectives we aim to achieve are as follows: 
% \subsection{Research Objectives} \label{researchobj}
The primary objective of this research is to investigate the performance of fine-tuned \emph{Whisper} models on accented speech and assess their generalization capabilities to other accents. Specifically, the study focuses on SEA-accented ATC speech, a domain that is underrepresented in publicly available datasets (see Table \ref{table:1}).

To achieve this, the following steps were undertaken:
\begin{enumerate}[label=\Alph*]
    \item \textbf{Dataset Creation:} Develop a SEA-accented ATC dataset, capturing the unique characteristics of the regional accent.
    \item \textbf{Model Fine-Tuning:} Fine-tune \emph{Whisper} models using the curated dataset and test data augmentation strategies to enhance transcription accuracy for Southeast Asian accents.
    \item \textbf{Model Benchmarking:} Benchmark the fine-tuned models against existing \emph{Whisper-ATC} models, such as those trained on ATCO2 and ATCOSIM datasets, which predominantly include Western-accented English.
    \item \textbf{Results Analysis:} The performance results were analyzed to identify improvements and limitations.
\end{enumerate}

In addition, this research addresses the practical limitations faced by military organizations in adopting ASR technologies. These include challenges in:
\begin{itemize}
    \item Securing access to diverse and high-quality datasets due to security and operational constraints \cite{Rashid2023ArtificialII}.
    \item Overcoming the high computational requirements for model training and inference, which are not feasible for resource-constrained environments \cite{Lu2023SecurityAP,HS2024HybridAU}.
\end{itemize}

To address these challenges, we propose a robust dataset creation pipeline and a tailored model selection process that balances performance with computational efficiency. Ultimately, this research aims to provide military organizations with a practical roadmap for fine-tuning ASR models to accommodate their local accents and operational needs.

%-------------------------------------------------------------------------------------------------------------------------
% Data Acquisition
%-------------------------------------------------------------------------------------------------------------------------

\begin{figure}[h!]
    \centering
    \includegraphics[width=0.20\textwidth]{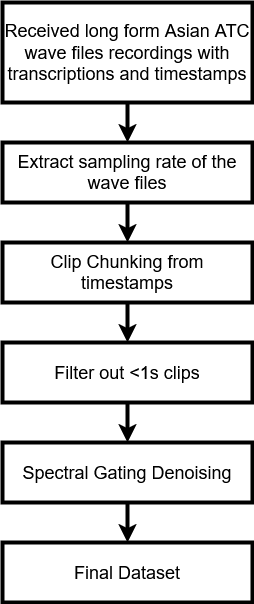}
    \caption{Dataset Creation Pipeline}
    \label{fig:data-pipeline}
\end{figure}

\subsection{Dataset Creation}
Our project obtained SEA-accented ATC data from an internal repository. If obtaining data from within the military may conflict with tight military security restrictions \cite{Lu2023SecurityAP}, Live-ATC\footnote{\url{https://www.liveatc.net/}}, which stores archived ATC communications from commercial airports all around the world, may be used instead. We then processed the dataset into a model-trainable format using the following pipeline as shown in Figure \ref{fig:data-pipeline}. 

\subsubsection*{Data Pipeline}
To enhance the quality of our dataset for fine-tuning, we used Spectral Gating denoising which operates by computing a spectrogram of an audio signal and estimating a noise threshold for each frequency band. This threshold is then used to create a mask that attenuates noise below the frequency-varying threshold, improving the clarity of the audio dataset \cite{sainburg2020finding}.

Data is separated into train, test, and validation splits with a proportion of 70\%, 15\%, and 15\% respectively after the ground-truth labels are obtained. The train split is used for model fine-tuning. The validation split is used to evaluate the performance of the fine-tuned model after each training epoch. The test split is used to benchmark model performance on data that is unseen by the model during the training process.

%-------------------------------------------------------------------------------------------------------------------------
% Model Finetuning
%-------------------------------------------------------------------------------------------------------------------------

\subsection{Model Fine-Tuning}

\emph{Whisper} offers a range of pre-trained model sizes for fine-tuning. Given the size, weight, and power (SWaP) constraints often faced in military computing~\cite{Dasari2019Complexity, Im2019Optimization}, our model selection prioritized both computational efficiency and accuracy. These models were fine-tuned on accent-specific ATC data, evaluated, and selected for testing.

\subsubsection{Model Selection}

\emph{Whisper} Small and \emph{Whisper} Large v3 Turbo were the selected models to fine-tune on the accented dataset. The reasons for our model selection are below.

\emph{Whisper} Small uses the traditional \emph{Whisper} architecture with 32 decoder layers \cite{RadfordWhisper2022}, balancing speed and accuracy. According to benchmarks\footnote{\url{https://github.com/SYSTRAN/faster-whisper/issues/1030}}\textsuperscript{,}\footnote{\url{https://paperswithcode.com/sota/automatic-speech-recognition-on-librispeech-8}}, \emph{Whisper} Small achieved 7.8\% WER on Librispeech benchmarks \cite{Librispeech}, comparable to \emph{Whisper} Medium at 7.432\% and 5.9\% with a significantly smaller model size and lesser training requirements. As \emph{Whisper} Tiny obtained a much higher WER of 17.15\%, model accuracy for \emph{Whisper} Tiny is heavily sacrificed for model speed, making it less suitable for use.

\emph{Whisper} Large v3 Turbo is a pruned version of \emph{Whisper} Large v3 with decoder layers cut down from 32 to 4 \cite{RadfordWhisper2022}. The same benchmarks show that the accuracy of \emph{Whisper} Large v3 Turbo is comparable to \emph{Whisper} Large v3 with significantly lesser compute requirements and faster inference. We thus decided to investigate the Turbo model's performance on ATC ASR. 
The model sizes comparison is shown in Table \ref{table:whispermodelcomparison}.

\begin{table}[h!]
\centering
\caption{\emph{Whisper} Model Comparison}

\begin{threeparttable}
\begin{tabular}{|l|c|c|c|}
\hline
\textbf{\emph{Whisper} Model} & \textbf{Parameters} & \textbf{Inference}$^{\nu}$ & \textbf{Training}$^{\nu}$\\ 
\hline
Tiny           & 39M   & 1GB   & 4GB   \\
Small          & 244M  & 2GB   & 8GB   \\
Medium         & 769M  & 5GB   & 20GB  \\
\hline
Large-v3       & 1550M & 10GB  & 40GB  \\
Large-v3 Turbo & 809M  & 6GB   & 24GB  \\
\hline
\end{tabular}
\begin{tablenotes}
\item $^{\nu}$Approximated VRAM requirements
\end{tablenotes}
\end{threeparttable}
\label{table:whispermodelcomparison}
\end{table}

\subsubsection{Augmentations}

To enhance the robustness of our fine-tuned Automatic Speech Recognition (ASR) models against noisy ATC environments, we introduced data augmentation techniques during the training process. By simulating various types of radio transmission noise, such as frequency filters and Hyperbolic Tangent distortion, we aimed to expose the model to a range of acoustic conditions typical of real-world ATC communications. This approach may help to allow the model to learn to distinguish between relevant speech signals and background noise, improving its ability to accurately transcribe speech even in suboptimal conditions. Augmentations are all available through the Audiomentations Library\footnote{\url{https://github.com/iver56/audiomentations}}.

\begin{itemize}
    \item \textbf{Frequency Filters.} Frequency filters include High Pass, Low Pass, and Band Pass filters. High Pass filters only allow high frequencies to be included in the signal, which could help reduce the low frequency signals such as mains harmonics. Low Pass filters only allow low frequencies to be included in the signal, which helps to remove White noise or Gaussian noise. Band Pass filters is a combination of High Pass and Low Pass filters, only allowing frequencies that fall within a certain range too be included in the signal. This helps to mimic the filtering properties of radio receivers, as well as exposing the model to different combinations of audio noise \cite{Boyer2023ReducingNA} \cite{Gardner2022ContinuousGW}.

    \item \textbf{Hyperbolic Tangent Distortion.} Hyperbolic Tangent Distortion ($\tanh$ distortion) is a symmetric distortion proportional to the
    loudness of the input and the pre-gain. Tanh distortion adds harmonics and changes the timbre of the sound, helping to model the nonlinearity of the radio receiver, mimicking radio noise \cite{Gardner2022ContinuousGW}.

\end{itemize}

\subsubsection{Model Fine-Tuning Results}

Model fine-tuning was conducted over the created dataset with and without augmentations on the train split over 30 epochs. After each training epoch, the model is evaluated on the validation set, and the combined WER on the validation set is calculated. For each model size, the checkpoint with the lowest validation WER was selected. The best configurations for each model are shown in Table \ref{table:modelFT}. 

% After conducting our training, the models with the best validation WER for \emph{Whisper Small} and \emph{Whisper Large v3 Turbo} is \textbf{0.122}, and \text{0.118} respectively.

%Best models
\begin{table}[h!]
\centering
\caption{Best fine-tuned \emph{Whisper} models}

\begin{threeparttable}
\begin{tabular}{|l|c|c|c|}
\hline
\textbf{Model Name} & \textbf{Augmentations} & \textbf{Validation WER}\\ 
\hline
sea-small          & Y & 0.122 \\
\hline
sea-large-v3-turbo & N & 0.118 \\ 
\hline
\end{tabular}
\end{threeparttable}
\label{table:modelFT}
\end{table}

%%%%%%%%%%%%%%%%%%%%%%%%%%%%%%%%%%%%%%%%%%%%%%%%%%%%%%%%%%%%%%%%%%%%%%%%%%%%%%%%%%%%%%%%%%%%%%%%%%%%%%%%%%%%%%%%%%%%%%%%%%
% EXPERIMENTAL RESULTS
%%%%%%%%%%%%%%%%%%%%%%%%%%%%%%%%%%%%%%%%%%%%%%%%%%%%%%%%%%%%%%%%%%%%%%%%%%%%%%%%%%%%%%%%%%%%%%%%%%%%%%%%%%%%%%%%%%%%%%%%%%
\section{Experimental Results} \label{results}

%-------------------------------------------------------------------------------------------------------------------------
% Model Assessment
%-------------------------------------------------------------------------------------------------------------------------

In order to evaluate the performance of our accent-specific fine-tuned model against currently available models \emph{Whisper-ATC}, cross inference tests were conducted on ATCOSIM and ATCO2 alongside our Southeast Asian-accented (SEA-accented) dataset. \emph{Whisper-ATC}'s models were trained on ATCOSIM (containing German, Swiss and French-accented English), and ATCO2 (containing less accented English ATC audio samples). This allows us to study the models' ability to generalize to different accents, and determine the effectiveness of our proposed fine-tuning methods on SEA-accented data.

\stepcounter{subsection}
\stepcounter{subsection}
\subsection{Model Benchmarking}
In automatic speech recognition (ASR), Word Error Rate (WER) is a critical metric for evaluating transcription accuracy. For high-quality, clean audio, state-of-the-art ASR systems have achieved WERs as low as benchmarks like Switchboard~\cite{Faria2022Switchboard}. However, in specialized domains such as air traffic control (ATC), achieving low WERs is more challenging due to factors like region-specific terminology, varied accents, and noisy communication channels. 

Recent studies in the ATC domain have reported WERs around 7.75\% across multiple databases~\cite{Zuluaga2020ATC}, suggesting that, while higher than those for clean audio, WERs below 25\% are attainable and may still be considered acceptable for certain transcription tasks~\cite{Munteanu2006Measuring,Fish2006WER}.

The best models were tested against currently available state-of-the-art models on the accent-specific datasets as well as datasets with accents separate from the domain of focus.

\begin{table}[h!]
\centering
\caption{model wer on ATCO2, ATCOSIM, SEA-accented datasets}
\begin{tabular}{@{}lccc@{}}
\toprule
\textbf{Model} & \textbf{ATCO2} & \textbf{ATCOSIM} & \textbf{SEA-accented}\\ \midrule
\multicolumn{4}{l}{\textbf{Small}}\\[2pt]
\raggedright openai/small & 0.9641 & 0.9106 & 1.2255 \\
\raggedright jlvdoorn/small-atco2-asr* & \textbf{0.4252} & 0.5432 & 0.7039 \\
\raggedright sea-small (ours) & 0.5735 & \textbf{0.5111} & \textbf{0.0982} \\
\midrule
\multicolumn{4}{l}{\textbf{Large v3}}\\[2pt]
\raggedright openai/large-v3 & 0.7896 & 0.8035 & 0.9340 \\
\raggedright jlvdoorn/large-v3-atco2-asr* & \textbf{0.3762} & \textbf{0.4625} & \textbf{0.5533} \\
\midrule
\multicolumn{4}{l}{\textbf{Large v3 Turbo}}\\[2pt]
\raggedright openai/large-v3-turbo & 0.8988 & 0.8171 & 1.1967 \\
\raggedright sea-large-v3-turbo (ours) & \textbf{0.5150} & \textbf{0.4110} & \textbf{0.1176} \\
\bottomrule
\end{tabular}
\begin{tablenotes}
\item * WHISPER-ATC models finetuned on ATCO2
\end{tablenotes}
\label{table:werresults}
\end{table}

Out of the \emph{Whisper} Small models, \textit{\emph{Whisper}-ATC's jlvdoorn/small-atco2-asr} model performed the best on ATCO2 with a WER of \textbf{0.4252}, while our fine-tuned model performed the best on ATCOSIM and our SEA-accented Dataset with \textbf{0.5111} and \textbf{0.0982} WER respectively.

Out of the \emph{Whisper} Large v3 models, \textit{\emph{Whisper}-ATC's jlvdoorn/large-v3-atco2-asr} model outperforms OpenAI's pretrained model on all datasets with \textbf{0.3762}, \textbf{0.4625} and \textbf{0.5533} WER for ATCO2, ATCOSIM and our SEA-accented Dataset respectively.

Out of the \emph{Whisper} Large v3 Turbo models, our fine-tuned model also outperforms OpenAI's pretrained model on all datasets with \textbf{0.5150}, \textbf{0.4110} and \textbf{0.1176} WER for ATCO2, ATCOSIM and our SEA-accented Dataset respectively.

On the SEA-accented Dataset, our fine-tuned \emph{Whisper} Small performed the bestwith \textbf{0.0982} WER. On ATCOSIM, our fine-tuned \emph{Whisper} Large v3 Turbo performed the best with \textbf{0.4110} WER. On ATCO2, \textit{\emph{Whisper}-ATC's jlvdoorn/large-v3-atco2-asr} performed the best with \textbf{0.3762} WER.

% Old one
% On the ATCO2 dataset, the best-performing model, \textit{jlvdoorn/large-v3-atco2-asr}, achieved a WER of \textbf{0.3762}. While this represents the strongest performance on this dataset, it remains above the desired threshold, reflecting the difficulty of transcribing noisy radio communications that rely on precise phraseology and waypoint references. Other models, including \textit{our large-v3-turbo} and \textit{our small}, also struggled, achieving WERs of \textbf{0.5150} and \textbf{0.5735}, respectively.

% For the ATCOSIM dataset, a closely related domain with slightly different recording conditions, the best WER was achieved by the \textit{our large-v3-turbo} model at \textbf{0.4110}. While slightly better than the performance on ATCO2, this result highlights the challenges of transcription in contexts where audio quality, phraseology, and jargon complexity vary. In such environments, the omission or substitution of key terms, such as waypoints or airport beacons, can disproportionately impact the transcription's utility.

% The SEA-accented dataset, representing accents and linguistic features distinct from the primary domain, exhibited additional challenges. Here, the best-performing model, \textit{our small}, achieved a WER of \textbf{0.0982}, indicating better handling of accents outside the aviation domain. However, most models showed significantly higher WERs, reflecting the difficulty of adapting to diverse linguistic characteristics in the absence of targeted fine-tuning.

\subsection{Results Analysis}

% The evaluated models demonstrated limited success across the datasets, with all WERs exceeding the 25\% threshold~\cite{Munteanu2006Measuring} for acceptable ATC transcription. Table~\ref{table:werresults} highlights the challenges of transcribing aviation-specific communications, including specialized terminology, non-standardized phraseology, and noisy radio conditions.

% Short audio clips further amplified these difficulties, as single errors disproportionately affected WER, underscoring the need for highly precise transcription. These results emphasize the importance of domain-specific fine-tuning and noise-robust training to address the unique demands of ATC transcription tasks. 

\subsubsection{Comparison of Model Performance}  

The results from \emph{Whisper} Large v3 and \emph{Whisper} Large v3 Turbo indicate that fine-tuning models on accented ATC data improves performance over their pretrained versions during inference. This improvement can be attributed to fine-tuning processes that make \emph{Whisper} models more robust to ATC-specific background noise, thereby enhancing transcription accuracy in this specialized domain.

The differences in model performance underscore the importance of training data quantity, quality, and specificity to the target domain. For example, while \textit{jlvdoorn/large-v3-atco2-asr} achieved a WER of \textbf{0.3762} on the ATCO2 dataset, it was trained on only 2 hours of ATCO-released data~\cite{Zuluaga_Gomez_2023}. Despite this limitation, its targeted fine-tuning enabled competitive performance on this specific dataset.

In contrast, our fine-tuned \textit{small} model achieved a significantly lower WER of \textbf{0.0982} on the SEA-accented dataset. This model benefited from 37 hours of region-specific training data, demonstrating the critical role of aligning training data with the test domain~\cite{Ircing2019ATCC}. However, on datasets with different accents and distinct lexicons, the same model experienced a sharp increase in WER, highlighting its inability to generalize effectively beyond its training domain~\cite{Wang2024ASR}. This issue is attributed to variations in accents, terminology, and phraseology. 

In ATC communications, standardized terminology is used to reference waypoints, beacons, and commands~\cite{ICAO2018Standardization}, which minimizes linguistic variance within a region but reduces adaptability to other contexts~\cite{FAA2025Glossary}. These findings indicate that training models on region-specific datasets leads to superior performance but limits generalization.

\subsubsection{Insights and Recommendations} 

Based on these results, several critical areas for improvement can be identified:
\begin{itemize}
    \item \textbf{Region-Specific Fine-Tuning.} Fine-tuning models on region-specific data is vital for addressing challenges such as specialized terminology and consistent noise patterns. While \textit{jlvdoorn/large-v3-atco2-asr} performed well on ATCO2, its performance on other datasets like ATCOSIM and the SEA-accented dataset highlights the need for broader adaptability.
    \item \textbf{Handling Noisy Inputs.} Noise-robust training techniques, such as data augmentation and spectral masking, are essential to mitigate the challenges posed by noisy ATC communications. Incorporating diverse noise profiles in training data could further improve robustness.
    \item \textbf{Accent Adaptation.} The disparity in WERs across datasets highlights the importance of accent-specific modeling. Strategies such as transfer learning, multilingual pretraining, and adaptive fine-tuning could enhance adaptability to diverse linguistic characteristics.
    \item \textbf{Evaluation Metrics for Short Transcripts.} In the ATC domain, where audio clips are often short, single errors disproportionately affect WER. Alternative metrics, such as per-word accuracy or semantic error rates, could provide a more nuanced understanding of model performance in this context.
\end{itemize}

\subsubsection{Broader Implications} 

The findings shed light on the trade-offs between generalization and region-specific accuracy in ASR systems. While general-purpose ASR models can be very accurate for regular speech, they struggle in specialized domains like ATC, where transcription errors can have significant consequences~\cite{Zuluaga2020ATC}. Integrating contextual knowledge, such as ATC-specific phraseology databases, into ASR models could further improve transcription accuracy and reduce WER.

The performance gap between current results and desired thresholds highlights the need for ongoing innovation in ASR modeling. Key areas include noise handling, region-specific adaptation, and accent robustness. Future research should focus on methods that balance generalization and specialization, such as accent-adaptive pretraining combined with region-specific fine-tuning~\cite{Zuluaga_Gomez_2023, Wang2024ASR}. Addressing these challenges will be critical for advancing the state-of-the-art in ATC transcription and ensuring the reliable deployment of ASR systems in safety-critical environments.

%%%%%%%%%%%%%%%%%%%%%%%%%%%%%%%%%%%%%%%%%%%%%%%%%%%%%%%%%%%%%%%%%%%%%%%%%%%%%%%%%%%%%%%%%%%%%%%%%%%%%%%%%%%%%%%%%%%%%%%%%%
% FUTURE WORK
%%%%%%%%%%%%%%%%%%%%%%%%%%%%%%%%%%%%%%%%%%%%%%%%%%%%%%%%%%%%%%%%%%%%%%%%%%%%%%%%%%%%%%%%%%%%%%%%%%%%%%%%%%%%%%%%%%%%%%%%%%
\section{Future Work} \label{futurework}

% \textcolor{red}{yz: please add some text to show we have 3 areas for future work...}
% For future work, we plan to find ways to further improve transcription accuracy in Air Traffic Control (ATC) Communications. 
Advancing Automatic Speech Recognition (ASR) technologies for Air Traffic Control (ATC) Communications remains an ongoing challenge, particularly in addressing the complexities of accented speech and noisy environments. In this work, we have demonstrated the potential of fine-tuned ASR models for accent-specific use cases. However, significant opportunities exist to further enhance transcription accuracy and model generalization. 

We have identified 3 key areas for further research: 1) incorporating dataset-specific contextual information into ASR models through prompting to improve region-specific transcription, 2) developing specialized denoising techniques tailored to the unique characteristics of ATC radio noise, and 3) advancing accent-agnostic ASR models capable of performing robustly across diverse linguistic and acoustic conditions. These directions aim to address both the technical limitations of current ASR systems and the practical constraints of real-world deployment in critical aviation domains.
% need to rephrase this ^^

\subsection{Addressing Region-Specific Terminology through Prompting}
Aviation communication relies heavily on region-specific terminology that varies significantly across geographic and regulatory contexts~\cite{Kowalski2022Normalization}. This variation poses a persistent challenge for transcription models. One potential solution to address unseen terminologies is the use of prompting, which may be able to guide model behavior in ASR tasks by providing contextual information prior to transcription. 

Currently, the best practices for utilizing prompts for Whisper are yet to be established, with current attempts demonstrating limited semantic understanding and accuracy improvements \cite{Yang2024Do}. Progress in Hugging Face's prompting methodologies has also been stagnant, with \texttt{prompt\_ids} failing to produce consistent improvements\footnote{https://github.com/huggingface/transformers/issues/23845}\textsuperscript{,}\footnote{\url{https://github.com/huggingface/transformers/pull/28687}}. 

Despite its promise, prompting strategies often demonstrate inconsistent performance gains in practice. This inconsistency highlights the need for refined methodologies to ensure reliable improvements in ASR tasks.

\subsection{ATC-specific Denoisers}
Our best fine-tuned model achieved only \textbf{0.0982} WER on our %created 
curated Southeast Asian-accented (SEA-accented) dataset, while pre-trained \emph{Whisper} models are able to perform to similar degrees of accuracy across different languages and accents without fine-tuning\cite{radford2023robust}. This elucidates a clear effect of ATC radio noise degrading \emph{Whisper} model performance. This is also to be expected as ATC noise is often lossy and characterized by low signal-to-noise ratios, which massively affects ASR performance\cite{Lin2021SpokenIU}. As a result, ATC-specific speech enhancers would need to be developed to be able to make up for the information lost. A notably successful attempt would be Recognition Oriented Speech Enhancement (ROSE) \cite{Yu2023ROSEAR}, which has been shown to improve ASR performance, although their models are not released to the public. Hence, there are grounds to further explore and experiment with developing ATC-specific denoisers to improve ASR in ATC.

% \subsection{LLM assisted ASR}

\subsection{Accent Generalization and Expansion}
Future research could focus on developing ASR models that robustly generalize across a broader spectrum of accents, particularly underrepresented ones such as those prevalent in East Asia. Incorporating additional accent-specific training data and leveraging self-supervised learning techniques may enhance the robustness of these systems in handling regional variations. Combining existing datasets with publicly available ones could further fine-tune models capable of generalizing to all accents. Building an overall system with improved performance will also require addressing noise robustness, as the impact of radio noise and environmental interference on transcription accuracy remains a critical challenge \cite{fritz2024analyzing,Iwamoto2022HowBA,Kumar2024PerformanceEO}. Investigating advanced noise reduction techniques and multi-modal approaches, such as combining visual or contextual data with audio inputs, could enhance transcription reliability in noisy environments.

The scarcity of publicly available, high-quality ATC datasets containing diverse accents presents a significant barrier to progress. Addressing this issue may involve data augmentation strategies, including simulated audio conditions, and encouraging the release of annotated datasets for non-commercial research purposes.

Addressing these challenges would help bridge the gap between current ASR capabilities and the requirements of real-world ATC operations, ultimately enhancing airspace safety and efficiency.

%%%%%%%%%%%%%%%%%%%%%%%%%%%%%%%%%%%%%%%%%%%%%%%%%%%%%%%%%%%%%%%%%%%%%%%%%%%%%%%%%%%%%%%%%%%%%%%%%%%%%%%%%%%%%%%%%%%%%%%%%%
% CONCLUSION
%%%%%%%%%%%%%%%%%%%%%%%%%%%%%%%%%%%%%%%%%%%%%%%%%%%%%%%%%%%%%%%%%%%%%%%%%%%%%%%%%%%%%%%%%%%%%%%%%%%%%%%%%%%%%%%%%%%%%%%%%%
\section{Conclusion} \label{conclusion}
% :> oops
Automatic Speech Recognition (ASR) technology holds significant promise for enhancing the operational efficiency and safety of aerial missions. However, existing ASR solutions face substantial challenges transcribing accented and noisy Air Traffic Control (ATC) communications, particularly in specialized and mission-critical domains. In this paper, we have demonstrated a replicable pipeline to fine-tune ASR models on accented ATC speech, leveraging pretrained generalist ASR models such as OpenAI's \emph{Whisper}. 

Our results highlight the importance of region-specific adaptation, as models fine-tuned on region-specific datasets significantly outperform their pretrained counterparts in transcription accuracy. By addressing challenges such as noise robustness, accent-specific modeling, and region-specific phraseology, our approach achieved substantial improvements in Word Error Rate (WER). These findings demonstrate the potential to develop effective ASR solutions for underrepresented accent groups. This is especially important for military organizations and similar stakeholders, where even minor transcription errors could compromise decision-making and operational safety.

Furthermore, the use of fine-tuning mitigates the catastrophic forgetting observed in previous efforts to train accent-robust ASR models from scratch. This approach democratizes access to accurate ASR systems for accent groups traditionally underrepresented in existing training corpora, bridging gaps in performance and accessibility. Future work should focus on improving generalization across domains and exploring multilingual and accent-adaptive pretraining strategies to further enhance the robustness of ASR systems for ATC and other specialized ASR applications.

%% lynus: im pretty sure this conclusion is missing something but im not sure what. maybe it just ends awkwardly?
%  Marcus: i referenced other similar papers and this is the conclusion i expanded from ur paragraph ^^ do share thoughts
%% it def feels more complete thanku

% to mention the near-real time applications also? from "Towards an accent-robust approach for ATC communications transcription", "(be able to verify what he/she heard on the radio by looking at the text transcription, be able to decipher non-native English accents from controllers, not lose time asking the ATC to repeat the message several times)." -> operational efficiency
% yz: advise to avoid mentioning, unless you intend to implement real-time, suggest future work

\section*{Acknowledgment} \label{acknowledgements}
This paper is made possible with the support of RSAF Agile Innovation Digital (RAiD), Republic of Singapore Air Force, Singapore.

\bibliographystyle{IEEEtran} % We choose the "IEEEtran" reference style
\bibliography{refs} % Entries are in the refs.bib file

\begin{thebibliography}{10}
\providecommand{\url}[1]{#1}
\csname url@samestyle\endcsname
\providecommand{\newblock}{\relax}
\providecommand{\bibinfo}[2]{#2}
\providecommand{\BIBentrySTDinterwordspacing}{\spaceskip=0pt\relax}
\providecommand{\BIBentryALTinterwordstretchfactor}{4}
\providecommand{\BIBentryALTinterwordspacing}{\spaceskip=\fontdimen2\font plus
\BIBentryALTinterwordstretchfactor\fontdimen3\font minus \fontdimen4\font\relax}
\providecommand{\BIBforeignlanguage}[2]{{%
\expandafter\ifx\csname l@#1\endcsname\relax
\typeout{** WARNING: IEEEtran.bst: No hyphenation pattern has been}%
\typeout{** loaded for the language `#1'. Using the pattern for}%
\typeout{** the default language instead.}%
\else
\language=\csname l@#1\endcsname
\fi
#2}}
\providecommand{\BIBdecl}{\relax}
\BIBdecl

\bibitem{Ohneiser2021IntegratingEA}
\BIBentryALTinterwordspacing
O.~Ohneiser, J.~Adamala, and I.-T. Salomea, ``Integrating eye- and mouse-tracking with assistant based speech recognition for interaction at controller working positions,'' \emph{Aerospace}, 2021. [Online]. Available: \url{https://api.semanticscholar.org/CorpusID:239250230}
\BIBentrySTDinterwordspacing

\bibitem{Fan2023EnhancingMS}
\BIBentryALTinterwordspacing
P.~Fan, D.~Guo, J.~Zhang, B.~Yang, and Y.~Lin, ``Enhancing multilingual speech recognition in air traffic control by sentence-level language identification,'' \emph{ArXiv}, vol. abs/2305.00170, 2023. [Online]. Available: \url{https://api.semanticscholar.org/CorpusID:258426556}
\BIBentrySTDinterwordspacing

\bibitem{Zuluaga2023LessonsLI}
\BIBentryALTinterwordspacing
J.~P. Zuluaga, I.~Nigmatulina, A.~Prasad, P.~Motl{\'i}cek, D.~Khalil, S.~R. Madikeri, A.~Tart, I.~Szoke, V.~Lenders, M.~Rigault, and K.~Choukri, ``{Lessons Learned in Transcribing 5000 h of Air Traffic Control Communications for Robust Automatic Speech Understanding},'' \emph{Aerospace}, 2023. [Online]. Available: \url{https://api.semanticscholar.org/CorpusID:264391192}
\BIBentrySTDinterwordspacing

\bibitem{radford2023robust}
A.~Radford, J.~W. Kim, T.~Xu, G.~Brockman, C.~McLeavey, and I.~Sutskever, ``Robust speech recognition via large-scale weak supervision,'' in \emph{International conference on machine learning}.\hskip 1em plus 0.5em minus 0.4em\relax PMLR, 2023, pp. 28\,492--28\,518.

\bibitem{Smith2020UnderstandingRA}
\BIBentryALTinterwordspacing
M.~Smith, M.~Strohmeier, V.~Lenders, and I.~Martinovic, ``{Understanding realistic attacks on airborne collision avoidance systems},'' \emph{Journal of Transportation Security}, vol.~15, pp. 87 -- 118, 2020. [Online]. Available: \url{https://api.semanticscholar.org/CorpusID:222124915}
\BIBentrySTDinterwordspacing

\bibitem{Zuluaga2020ATC}
\BIBentryALTinterwordspacing
J.~Zuluaga-Gomez, P.~Motlicek, Q.~Zhan, K.~Vesel{\'y}, and R.~Braun, ``{Automatic Speech Recognition Benchmark for Air-Traffic Communications},'' in \emph{Interspeech 2020}.\hskip 1em plus 0.5em minus 0.4em\relax ISCA, Oct. 2020, pp. 2297--2301. [Online]. Available: \url{http://dx.doi.org/10.21437/interspeech.2020-2173}
\BIBentrySTDinterwordspacing

\bibitem{Wang2024ASR}
\BIBentryALTinterwordspacing
Z.~Wang, P.~Jiang, Z.~Wang, B.~Han, H.~Liang, Y.~Ai, and W.~Pan, ``{Enhancing Air Traffic Control Communication Systems with Integrated Automatic Speech Recognition: Models, Applications and Performance Evaluation},'' \emph{Sensors}, vol.~24, no.~14, p. 4715, 2024. [Online]. Available: \url{https://www.mdpi.com/1424-8220/24/14/4715}
\BIBentrySTDinterwordspacing

\bibitem{moon2011air}
W.-C. Moon, K.-E. Yoo, Y.-C. Choi \emph{et~al.}, ``Air traffic volume and air traffic control human errors,'' \emph{Journal of Transportation Technologies}, vol.~1, no.~03, p.~47, 2011.

\bibitem{Zdorova2021DoWR}
\BIBentryALTinterwordspacing
N.~Zdorova, S.~Malyutina, A.~K. Laurinavichyute, A.~Kaprielova, A.~Ziubanova, and A.~Lopukhina, ``Do we rely on good-enough processing in reading under auditory and visual noise?'' \emph{PLOS ONE}, vol.~18, 2021. [Online]. Available: \url{https://api.semanticscholar.org/CorpusID:239420488}
\BIBentrySTDinterwordspacing

\bibitem{kopald2013applying}
H.~D. Kopald, A.~Chanen, S.~Chen, E.~C. Smith, and R.~M. Tarakan, ``Applying automatic speech recognition technology to air traffic management,'' in \emph{2013 IEEE/AIAA 32nd Digital Avionics Systems Conference (DASC)}.\hskip 1em plus 0.5em minus 0.4em\relax IEEE, 2013, pp. 6C3--1.

\bibitem{Wang2024EnhancingAT}
\BIBentryALTinterwordspacing
Z.~Wang, P.~Jiang, Z.~Wang, B.~Han, H.~Liang, Y.~Ai, and W.~Pan, ``{Enhancing Air Traffic Control Communication Systems with Integrated Automatic Speech Recognition: Models, Applications and Performance Evaluation},'' \emph{Sensors (Basel, Switzerland)}, vol.~24, 2024. [Online]. Available: \url{https://api.semanticscholar.org/CorpusID:271392005}
\BIBentrySTDinterwordspacing

\bibitem{LDC94S14A}
\BIBentryALTinterwordspacing
J.~J. Godfrey, ``Air traffic control complete (atcc),'' 1994, lDC94S14A. [Online]. Available: \url{https://catalog.ldc.upenn.edu/LDC94S14A}
\BIBentrySTDinterwordspacing

\bibitem{segura2007hiwire}
J.~Segura, T.~Ehrette, A.~Potamianos, D.~Fohr, I.~Illina, P.-A. Breton, V.~Clot, R.~Gemello, M.~Matassoni, and P.~Maragos, ``The hiwire database: A noisy and non-native english speech corpus for cockpit communication,'' in \emph{Proceedings of the 8th Annual Conference of the International Speech Communication Association (INTERSPEECH 2007)}, 2007, pp. 1493--1496.

\bibitem{hofbauer2008atcosim}
K.~Hofbauer, S.~Petrik, and H.~Hering, ``{The ATCOSIM Corpus of Non-Prompted Clean Air Traffic Control Speech},'' in \emph{LREC}.\hskip 1em plus 0.5em minus 0.4em\relax Citeseer, 2008.

\bibitem{zuluaga2023atco2}
\BIBentryALTinterwordspacing
J.~Zuluaga-Gomez, K.~Veselý, I.~Szöke, A.~Blatt, P.~Motlicek, M.~Kocour, M.~Rigault, K.~Choukri, A.~Prasad, S.~S. Sarfjoo, I.~Nigmatulina, C.~Cevenini, P.~Kolčárek, A.~Tart, J.~Černocký, and D.~Klakow, ``{ATCO2 corpus: A Large-Scale Dataset for Research on Automatic Speech Recognition and Natural Language Understanding of Air Traffic Control Communications},'' 2023. [Online]. Available: \url{https://arxiv.org/abs/2211.04054}
\BIBentrySTDinterwordspacing

\bibitem{yang2019atcspeech}
\BIBentryALTinterwordspacing
B.~Yang, X.~Tan, Z.~Chen, B.~Wang, D.~Li, Z.~Yang, X.~Wu, and Y.~Lin, ``{ATCSpeech: a multilingual pilot-controller speech corpus from real Air Traffic Control environment},'' \emph{CoRR}, vol. abs/1911.11365, 2019. [Online]. Available: \url{http://arxiv.org/abs/1911.11365}
\BIBentrySTDinterwordspacing

\bibitem{hofbauer-etal-2008-atcosim}
\BIBentryALTinterwordspacing
K.~Hofbauer, S.~Petrik, and H.~Hering, ``{The {ATCOSIM} Corpus of Non-Prompted Clean Air Traffic Control Speech},'' in \emph{Proceedings of the Sixth International Conference on Language Resources and Evaluation ({LREC}`08)}, N.~Calzolari, K.~Choukri, B.~Maegaard, J.~Mariani, J.~Odijk, S.~Piperidis, and D.~Tapias, Eds.\hskip 1em plus 0.5em minus 0.4em\relax Marrakech, Morocco: European Language Resources Association (ELRA), May 2008. [Online]. Available: \url{https://aclanthology.org/L08-1507/}
\BIBentrySTDinterwordspacing

\bibitem{rataj2019aclistant}
J.~Rataj, H.~Helmke, and O.~Ohneiser, ``{AcListant} with {Continuous} {Learning}: {Speech} {Recognition} in {Air} {Traffic} {Control},'' in \emph{EIWAC 2019}, Oct. 2019.

\bibitem{pellegrini2018airbus}
\BIBentryALTinterwordspacing
T.~Pellegrini, J.~Farinas, E.~Delpech, and F.~Lancelot, ``{The Airbus Air Traffic Control speech recognition 2018 challenge: towards {ATC} automatic transcription and call sign detection},'' \emph{CoRR}, vol. abs/1810.12614, 2018. [Online]. Available: \url{http://arxiv.org/abs/1810.12614}
\BIBentrySTDinterwordspacing

\bibitem{smidl2019ATCC}
L.~Šmídl, J.~Švec, D.~Tihelka, J.~Matoušek, J.~Romportl, and P.~Ircing, ``{Air traffic control communication (ATCC) speech corpora and their use for ASR and TTS development},'' \emph{Language Resources and Evaluation}, vol.~53, 02 2019.

\bibitem{helmke2023haawaii}
H.~Helmke, M.~Kleinert, A.~Linß, L.~K. Petr~Motlicek, Hanno~Wiese, J.~Harfmann, H.~A. Nuno~Cebola, and T.~Simiganoschi, ``{The HAAWAII Framework for Automatic SpeechUnderstanding of Air Traffic Communication},'' 11 2023.

\bibitem{Pan2024AssessmentAA}
\BIBentryALTinterwordspacing
W.~Pan, J.~Zhang, Y.~Zhang, P.~Jiang, and S.~Han, ``Assessment and analysis of accents in air traffic control speech: a fusion of deep learning and information theory,'' \emph{Frontiers in Neurorobotics}, vol.~18, 2024. [Online]. Available: \url{https://api.semanticscholar.org/CorpusID:268533437}
\BIBentrySTDinterwordspacing

\bibitem{Kheddar2023DeepTL}
\BIBentryALTinterwordspacing
H.~Kheddar, Y.~Himeur, S.~A. Al-Maadeed, A.~Amira, and F.~Bensaali, ``Deep transfer learning for automatic speech recognition: Towards better generalization,'' \emph{ArXiv}, vol. abs/2304.14535, 2023. [Online]. Available: \url{https://api.semanticscholar.org/CorpusID:258418066}
\BIBentrySTDinterwordspacing

\bibitem{Ahrenhold2023ValidatingAS}
\BIBentryALTinterwordspacing
N.~Ahrenhold, H.~Helmke, T.~M{\"u}hlhausen, O.~Ohneiser, M.~Kleinert, H.~Ehr, L.~Klamert, and J.~P. Zuluaga, ``{Validating Automatic Speech Recognition and Understanding for Pre-Filling Radar Labels—Increasing Safety While Reducing Air Traffic Controllers’ Workload},'' \emph{Aerospace}, 2023. [Online]. Available: \url{https://api.semanticscholar.org/CorpusID:259115093}
\BIBentrySTDinterwordspacing

\bibitem{Iwamoto2022HowBA}
\BIBentryALTinterwordspacing
K.~Iwamoto, T.~Ochiai, M.~Delcroix, R.~Ikeshita, H.~Sato, S.~Araki, and S.~Katagiri, ``{How Bad Are Artifacts?: Analyzing the Impact of Speech Enhancement Errors on ASR},'' in \emph{Interspeech}, 2022. [Online]. Available: \url{https://api.semanticscholar.org/CorpusID:246015341}
\BIBentrySTDinterwordspacing

\bibitem{Prabhu2023AccentedSR}
\BIBentryALTinterwordspacing
D.~Prabhu, P.~Jyothi, S.~Ganapathy, and V.~Unni, ``{Accented Speech Recognition With Accent-specific Codebooks},'' \emph{ArXiv}, vol. abs/2310.15970, 2023. [Online]. Available: \url{https://api.semanticscholar.org/CorpusID:264439549}
\BIBentrySTDinterwordspacing

\bibitem{li2021accent}
J.~Li, V.~Manohar, P.~Chitkara, A.~Tjandra, M.~Picheny, F.~Zhang, X.~Zhang, and Y.~Saraf, ``Accent-robust automatic speech recognition using supervised and unsupervised wav2vec embeddings,'' \emph{arXiv preprint arXiv:2110.03520}, 2021.

\bibitem{Deng2021ImprovingAI}
\BIBentryALTinterwordspacing
K.~Deng, S.~Cao, and L.~Ma, ``{Improving Accent Identification and Accented Speech Recognition Under a Framework of Self-supervised Learning},'' \emph{ArXiv}, vol. abs/2109.07349, 2021. [Online]. Available: \url{https://api.semanticscholar.org/CorpusID:237513397}
\BIBentrySTDinterwordspacing

\bibitem{jahchan2021towards}
N.~Jahchan, F.~Barbier, A.~D. Gita, K.~Khelif, and E.~Delpech, ``{Towards an Accent-Robust Approach for ATC Communications Transcription.}'' in \emph{Interspeech}, 2021, pp. 3281--3285.

\bibitem{Jelassi2024RevolutionizingRA}
\BIBentryALTinterwordspacing
M.~Jelassi, O.~Jemai, and J.~Demongeot, ``{Revolutionizing Radiological Analysis: The Future of French Language Automatic Speech Recognition in Healthcare},'' \emph{Diagnostics}, vol.~14, 2024. [Online]. Available: \url{https://api.semanticscholar.org/CorpusID:269421510}
\BIBentrySTDinterwordspacing

\bibitem{Graham2024EvaluatingOW}
\BIBentryALTinterwordspacing
C.~Graham and N.~Roll, ``{Evaluating OpenAI's Whisper ASR: Performance analysis across diverse accents and speaker traits},'' \emph{JASA Express Letters}, vol.~4, no.~2, p. 025206, 02 2024. [Online]. Available: \url{https://doi.org/10.1121/10.0024876}
\BIBentrySTDinterwordspacing

\bibitem{Sanabria2023TheEI}
\BIBentryALTinterwordspacing
R.~Sanabria, N.~Bogoychev, N.~Markl, A.~Carmantini, O.~Klejch, and P.~Bell, ``{The Edinburgh International Accents of English Corpus: Towards the Democratization of English ASR},'' \emph{ICASSP 2023 - 2023 IEEE International Conference on Acoustics, Speech and Signal Processing (ICASSP)}, pp. 1--5, 2023. [Online]. Available: \url{https://api.semanticscholar.org/CorpusID:257901049}
\BIBentrySTDinterwordspacing

\bibitem{pan2024assessment}
W.~Pan, J.~Zhang, Y.~Zhang, P.~Jiang, and S.~Han, ``Assessment and analysis of accents in air traffic control speech: a fusion of deep learning and information theory,'' \emph{Frontiers in neurorobotics}, vol.~18, p. 1360094, 2024.

\bibitem{han2021supervised}
T.~Han, H.~Huang, Z.~Yang, and W.~Han, ``Supervised contrastive learning for accented speech recognition,'' \emph{arXiv preprint arXiv:2107.00921}, 2021.

\bibitem{aksenova2022AccentedSR}
\BIBentryALTinterwordspacing
A.~Aksenova, Z.~Chen, C.-C. Chiu, D.~van Esch, P.~Golik, W.~Han, L.~King, B.~Ramabhadran, A.~Rosenberg, S.~Schwartz, and G.~Wang, ``Accented speech recognition: Benchmarking, pre-training, and diverse data,'' \emph{ArXiv}, vol. abs/2205.08014, 2022. [Online]. Available: \url{https://api.semanticscholar.org/CorpusID:248834135}
\BIBentrySTDinterwordspacing

\bibitem{Maison2023ImprovingAS}
\BIBentryALTinterwordspacing
L.~Maison and Y.~Est{\`e}ve, ``{Improving Accented Speech Recognition with Multi-Domain Training},'' \emph{ICASSP 2023 - 2023 IEEE International Conference on Acoustics, Speech and Signal Processing (ICASSP)}, pp. 1--5, 2023. [Online]. Available: \url{https://api.semanticscholar.org/CorpusID:257505352}
\BIBentrySTDinterwordspacing

\bibitem{RadfordWhisper2022}
\BIBentryALTinterwordspacing
A.~Radford, J.~W. Kim, T.~Xu, G.~Brockman, C.~McLeavey, and I.~Sutskever, ``{Robust Speech Recognition via Large-Scale Weak Supervision},'' 2022. [Online]. Available: \url{https://arxiv.org/abs/2212.04356}
\BIBentrySTDinterwordspacing

\bibitem{Lin2021SpokenIU}
\BIBentryALTinterwordspacing
Y.~Lin, ``{Spoken Instruction Understanding in Air Traffic Control: Challenge, Technique, and Application},'' \emph{Aerospace}, 2021. [Online]. Available: \url{https://api.semanticscholar.org/CorpusID:233789045}
\BIBentrySTDinterwordspacing

\bibitem{Yu2023ROSEAR}
\BIBentryALTinterwordspacing
X.~Yu, D.~Guo, J.~Zhang, and Y.~Lin, ``{ROSE: A Recognition-Oriented Speech Enhancement Framework in Air Traffic Control Using Multi-Objective Learning},'' \emph{IEEE/ACM Transactions on Audio, Speech, and Language Processing}, vol.~32, pp. 3365--3378, 2023. [Online]. Available: \url{https://api.semanticscholar.org/CorpusID:266162865}
\BIBentrySTDinterwordspacing

\bibitem{zeek1949investigation}
R.~W. Zeek, ``{Investigation and Analysis of "Capture Effect" in F-M and A-M Communication Systems},'' Naval Research Laboratory, Washington, DC, Tech. Rep., 1949.

\bibitem{fritz2024analyzing}
F.~Fritz, A.~Cornaggia-Urrigshardt, L.~Henneke, F.~Kurth, and K.~Wilkinghoff, ``{Analyzing the Impact of HF-Specific Signal Degradation on Automatic Speech Recognition},'' in \emph{2024 International Conference on Military Communication and Information Systems (ICMCIS)}.\hskip 1em plus 0.5em minus 0.4em\relax IEEE, 2024, pp. 1--10.

\bibitem{Chen2023EffectsOL}
\BIBentryALTinterwordspacing
S.~Chen, H.~Helmke, R.~M. Tarakan, O.~Ohneiser, H.~D. Kopald, and M.~Kleinert, ``{Effects of Language Ontology on Transatlantic Automatic Speech Understanding Research Collaboration in the Air Traffic Management Domain},'' \emph{Aerospace}, 2023. [Online]. Available: \url{https://api.semanticscholar.org/CorpusID:259053606}
\BIBentrySTDinterwordspacing

\bibitem{jlvdoorn2024WhisperATC}
J.~van Doorn, J.~Sun, J.~Hoekstra, P.~Jonk, and V.~de~Vries, ``{Whisper-ATC Open Models for Air Traffic Control Automatic Speech Recognition with Accuracy},'' \emph{ICRAT}, 2024.

\bibitem{Xie2024FromFE}
\BIBentryALTinterwordspacing
X.~Xie and C.~Kurumada, ``From first encounters to longitudinal exposure: a repeated exposure-test paradigm for monitoring speech adaptation,'' \emph{Frontiers in Psychology}, vol.~15, 2024. [Online]. Available: \url{https://api.semanticscholar.org/CorpusID:270165369}
\BIBentrySTDinterwordspacing

\bibitem{Romanov2021TheUM}
\BIBentryALTinterwordspacing
A.~S. Romanov, S.~A. Stepanov, M.~V. Poluboyarova, M.~Angaleva, and N.~Belaya, ``The u.s. military culture system of values mirrored by professional lingo,'' \emph{SHS Web of Conferences}, 2021. [Online]. Available: \url{https://api.semanticscholar.org/CorpusID:235836591}
\BIBentrySTDinterwordspacing

\bibitem{sun2018domain}
S.~Sun, C.-F. Yeh, M.-Y. Hwang, M.~Ostendorf, and L.~Xie, ``Domain adversarial training for accented speech recognition,'' in \emph{2018 IEEE international conference on acoustics, speech and signal processing (ICASSP)}.\hskip 1em plus 0.5em minus 0.4em\relax IEEE, 2018, pp. 4854--4858.

\bibitem{Rashid2023ArtificialII}
\BIBentryALTinterwordspacing
A.~B. Rashid, A.~K. Kausik, A.~A.~H. Sunny, and M.~H. Bappy, ``{Artificial Intelligence in the Military: An Overview of the Capabilities, Applications, and Challenges},'' \emph{Int. J. Intell. Syst.}, vol. 2023, pp. 1--31, 2023. [Online]. Available: \url{https://api.semanticscholar.org/CorpusID:265200275}
\BIBentrySTDinterwordspacing

\bibitem{Lu2023SecurityAP}
\BIBentryALTinterwordspacing
Y.~Lu, ``{Security and Privacy of Internet of Things: A Review of Challenges and Solutions},'' \emph{J. Cyber Secur. Mobil.}, vol.~12, pp. 813--844, 2023. [Online]. Available: \url{https://api.semanticscholar.org/CorpusID:265287997}
\BIBentrySTDinterwordspacing

\bibitem{HS2024HybridAU}
\BIBentryALTinterwordspacing
H.~H. S and J.~Nagaraja, ``{Hybrid Approach Using Machine Learning and IOT for Soldier Rescue : A Review},'' \emph{International Journal of Innovative Science and Research Technology (IJISRT)}, 2024. [Online]. Available: \url{https://api.semanticscholar.org/CorpusID:270991642}
\BIBentrySTDinterwordspacing

\bibitem{sainburg2020finding}
T.~Sainburg, M.~Thielk, and T.~Q. Gentner, ``Finding, visualizing, and quantifying latent structure across diverse animal vocal repertoires,'' \emph{PLoS computational biology}, vol.~16, no.~10, p. e1008228, 2020.

\bibitem{Dasari2019Complexity}
\BIBentryALTinterwordspacing
V.~R. Dasari, M.~S. Im, and B.~Geerhart, ``Complexity and mission computability of adaptive computing systems,'' \emph{The Journal of Defense Modeling and Simulation: Applications, Methodology, Technology}, vol.~19, no.~1, pp. 5--11, sep 2019. [Online]. Available: \url{http://dx.doi.org/10.1177}
\BIBentrySTDinterwordspacing

\bibitem{Im2019Optimization}
\BIBentryALTinterwordspacing
M.~S. Im, V.~R. Dasari, L.~Beshaj, and D.~R. Shires, ``Optimization problems with low swap tactical computing,'' in \emph{Disruptive Technologies in Information Sciences II}, M.~Blowers, R.~D. Hall, and V.~R. Dasari, Eds.\hskip 1em plus 0.5em minus 0.4em\relax SPIE, may 2019, p.~14. [Online]. Available: \url{http://dx.doi.org/10.1117/12.2518917}
\BIBentrySTDinterwordspacing

\bibitem{Librispeech}
V.~Panayotov, G.~Chen, D.~Povey, and S.~Khudanpur, ``{Librispeech: An ASR corpus based on public domain audio books},'' in \emph{2015 IEEE International Conference on Acoustics, Speech and Signal Processing (ICASSP)}, 2015, pp. 5206--5210.

\bibitem{Boyer2023ReducingNA}
\BIBentryALTinterwordspacing
M.~Boyer, L.~J. Bouyer, J.~S. Roy, and A.~Campeau-Lecours, ``{Reducing Noise, Artifacts and Interference in Single-Channel EMG Signals: A Review},'' \emph{Sensors (Basel, Switzerland)}, vol.~23, 2023. [Online]. Available: \url{https://api.semanticscholar.org/CorpusID:257481291}
\BIBentrySTDinterwordspacing

\bibitem{Gardner2022ContinuousGW}
\BIBentryALTinterwordspacing
J.~W. Gardner, H.~Middleton, C.~Liu, A.~Melatos, R.~J. Evans, W.~Moran, D.~Beniwal, H.~T. Cao, C.~Ingram, D.~Brown, and S.~W.~S. Ng, ``Continuous gravitational waves in the lab: recovering audio signals with a table-top optical microphone,'' 2022. [Online]. Available: \url{https://api.semanticscholar.org/CorpusID:245853636}
\BIBentrySTDinterwordspacing

\bibitem{Faria2022Switchboard}
\BIBentryALTinterwordspacing
A.~Faria, A.~Janin, S.~Adkoli, and K.~Riedhammer, ``{Toward Zero Oracle Word Error Rate on the Switchboard Benchmark},'' in \emph{Interspeech 2022}.\hskip 1em plus 0.5em minus 0.4em\relax ISCA, Sep. 2022, pp. 3973--3977. [Online]. Available: \url{http://dx.doi.org/10.21437/interspeech.2022-10959}
\BIBentrySTDinterwordspacing

\bibitem{Munteanu2006Measuring}
\BIBentryALTinterwordspacing
C.~Munteanu, G.~Penn, R.~Baecker, E.~Toms, and D.~James, ``Measuring the acceptable word error rate of machine-generated webcast transcripts,'' in \emph{Interspeech 2006}.\hskip 1em plus 0.5em minus 0.4em\relax ISCA, Sep. 2006, pp. 1756--Mon1CaP.2--0. [Online]. Available: \url{http://dx.doi.org/10.21437/interspeech.2006-40}
\BIBentrySTDinterwordspacing

\bibitem{Fish2006WER}
\BIBentryALTinterwordspacing
R.~Fish, Q.~Hu, and S.~Boykin, ``{Using Audio Quality to Predict Word Error Rate in an Automatic Speech Recognition System},'' The MITRE Corporation, Technical Report, 2006. [Online]. Available: \url{https://www.mitre.org/sites/default/files/pdf/06_1154.pdf}
\BIBentrySTDinterwordspacing

\bibitem{Zuluaga_Gomez_2023}
\BIBentryALTinterwordspacing
J.~Zuluaga-Gomez, I.~Nigmatulina, A.~Prasad, P.~Motlicek, D.~Khalil, S.~Madikeri, A.~Tart, I.~Szoke, V.~Lenders, M.~Rigault, and K.~Choukri, ``{Lessons Learned in Transcribing 5000 h of Air Traffic Control Communications for Robust Automatic Speech Understanding},'' \emph{Aerospace}, vol.~10, no.~10, p. 898, oct 2023. [Online]. Available: \url{http://dx.doi.org/10.3390/aerospace10100898}
\BIBentrySTDinterwordspacing

\bibitem{Ircing2019ATCC}
\BIBentryALTinterwordspacing
P.~Ircing and D.~Tihelka, ``Air traffic control communication (atcc) speech corpora and their use for asr and tts development,'' \emph{Language Resources and Evaluation}, vol.~53, pp. 601--621, 2019. [Online]. Available: \url{https://link.springer.com/article/10.1007/s10579-019-09449-5}
\BIBentrySTDinterwordspacing

\bibitem{ICAO2018Standardization}
\BIBentryALTinterwordspacing
{International Civil Aviation Organization (ICAO)}, \emph{Manual of Radiotelephony}, 2018, accessed: January 8, 2025. [Online]. Available: \url{https://skybrary.aero/sites/default/files/bookshelf/115.pdf}
\BIBentrySTDinterwordspacing

\bibitem{FAA2025Glossary}
\BIBentryALTinterwordspacing
{Federal Aviation Administration (FAA)}, \emph{Pilot/Controller Glossary}, 2025, accessed: January 8, 2025. [Online]. Available: \url{https://www.faa.gov/air_traffic/publications/atpubs/pcg_html/}
\BIBentrySTDinterwordspacing

\bibitem{Kowalski2022Normalization}
\BIBentryALTinterwordspacing
M.~Kowalski, P.~Kaczmarek, and B.~Kostek, ``Normalization of audio signals for the needs of machine learning,'' in \emph{2022 Signal Processing: Algorithms, Architectures, Arrangements, and Applications (SPA)}.\hskip 1em plus 0.5em minus 0.4em\relax IEEE, 2022, pp. 1--6. [Online]. Available: \url{https://ieeexplore.ieee.org/document/10372705}
\BIBentrySTDinterwordspacing

\bibitem{Yang2024Do}
\BIBentryALTinterwordspacing
C.-K. Yang, K.-P. Huang, and H.-y. Lee, ``{Do Prompts Really Prompt? Exploring the Prompt Understanding Capability of Whisper},'' \emph{arXiv preprint arXiv:2406.05806}, 2024. [Online]. Available: \url{https://arxiv.org/abs/2406.05806}
\BIBentrySTDinterwordspacing

\bibitem{Kumar2024PerformanceEO}
\BIBentryALTinterwordspacing
S.~Kumar, I.~Thorbecke, S.~Burdisso, E.~Villatoro-Tello, E.~ManjunathK, K.~Haciouglu, P.~Rangappa, P.~Motlicek, A.~Ganapathiraju, and A.~Stolcke, ``Performance evaluation of slam-asr: The good, the bad, the ugly, and the way forward,'' in \emph{ICASSP 2025 SALMA Workshop}, 2024. [Online]. Available: \url{https://api.semanticscholar.org/CorpusID:273850605}
\BIBentrySTDinterwordspacing

\end{thebibliography}
% references.bib loads from zotero, but it doesn't contain all citations!

\end{document}